\documentclass[journal]{IEEEtran}
\IEEEoverridecommandlockouts
\usepackage{cite}
\usepackage{amsmath,amssymb,amsfonts}
\usepackage{graphicx}
\usepackage{textcomp}
\usepackage{xcolor}
\usepackage{bm}
\usepackage{amsthm}
\usepackage{mathrsfs}
\usepackage{enumerate}
\usepackage{multirow}
\usepackage{color}
\usepackage{threeparttable}
\usepackage{subfigure}
\usepackage{booktabs}
\usepackage[pagebackref=false,breaklinks=true,colorlinks,bookmarks=false]{hyperref}
\usepackage{times}
\usepackage{array}
\usepackage{verbatim}
\usepackage{algorithm}
\usepackage{bbding}
\usepackage{algpseudocode}
\usepackage{lettrine}
\usepackage{placeins}
\usepackage{pifont}
\def\BibTeX{{\rm B\kern-.05em{\sc i\kern-.025em b}\kern-.08em
    T\kern-.1667em\lower.7ex\hbox{E}\kern-.125emX}}
\begin{document}
\title{IB-HFN: Information Bottleneck-Driven SAR-Optical Fusion Network for High-Fidelity Cloud Removal}
\author{Haojun Guo, Fan Feng, Ziquan Wang, Yongsheng Zhang, and Ying Yu% <-this % stops a space
\thanks{This work was supported in part by the National Natural Science Foundation of China under Grant 42071340, and in part by the Songshan Laboratory project (into the Henan Province major science and technology special management system) under Grant 221100211000-01.}% <-this % stops a space
\thanks{Haojun Guo, Fan Feng, Ziquan Wang, Yongsheng Zhang, and Ying Yu are with the Institute of Geospatial Information, Information Engineering University, Zhengzhou 450001, China (e-mail: ghjstc2012@163.com; fengrs1991@163.com; aresdrw@163.com; yszhang2001@vip.163.com; yuying5559104@163.com).}% <-this % stops a space
\thanks{Corresponding author: Ying Yu.}
}
\maketitle
\begin{abstract}
Synthetic aperture radar (SAR)-assisted optical cloud removal aims to recover surface information obscured by clouds in optical remote sensing images by exploiting complementary SAR observations. Existing multimodal fusion methods typically rely on direct spatial concatenation and pixel-wise supervision, which can propagate SAR speckle noise into optical reconstruction and lead to over-smoothed results. To address these limitations, we propose an Information Bottleneck-driven High-Fidelity Network (IB-HFN) for SAR-assisted optical cloud removal. IB-HFN employs a dual-stream backbone to preserve modality-specific representations before deep semantic fusion, thereby mitigating premature cross-modal contamination. At the fusion stage, we introduce a Spatial Information Bottleneck Fusion module that compresses SAR features through a channel-wise variational information bottleneck to suppress unstructured speckle noise. In parallel, a local-global gating mechanism predicts clear-sky regions and routes reliable optical details through a Dirac-initialized skip connection, decoupling noise suppression from texture preservation. We further develop a joint optimization strategy that integrates feature-level bottleneck regularization with image-level constraints on reconstruction accuracy, structural consistency, spectral fidelity, and contrastive sharpness. A dynamic weighting schedule balances these objectives to stabilize training and reduce hazy artifacts. Experiments on the SEN12MS-CR dataset under challenging spatio-temporal splits demonstrate that IB-HFN achieves superior structural preservation and spectral fidelity over existing methods.
\end{abstract}

\section{Introduction}
\IEEEPARstart{W}{ith} the rapid development of Earth observation technologies, high-resolution multispectral optical remote sensing imagery has become essential for a wide range of applications, including global land-cover classification, agricultural monitoring, and disaster assessment. However, more than half of annually acquired land-surface optical satellite images are affected by cloud contamination, leading to substantial loss of surface information~\cite{king2013spatial}. To mitigate this limitation, synthetic aperture radar (SAR) has been widely introduced to assist optical cloud removal because it actively measures microwave backscatter and can penetrate cloud cover. Unlike passive optical sensors, SAR provides complementary geometric and dielectric priors of the Earth's surface, enabling deep fusion models to infer missing optical content from SAR observations~\cite{zhang2024cloudguided,wang2024transfusion}. Nevertheless, the substantial imaging gap between active microwave sensing and passive optical imaging makes SAR-optical fusion highly challenging. Directly combining the two modalities may propagate SAR speckle noise into the optical reconstruction process, resulting in spectral distortion and degraded texture fidelity. Therefore, a key challenge in multimodal cloud removal is to suppress modality-specific noise, including optical cloud residuals and SAR speckles, while preserving high-frequency spatial structures and spectral consistency~\cite{liu2024structure}.

Existing multimodal cloud removal methods can be broadly categorized into two groups. The first group performs early data-level fusion or simple spatial feature concatenation. Classical methods exploit statistical priors, such as class-based linear regression with iterative residual compensation, to recover missing information from temporal observations~\cite{guo2026clear}. With the development of deep learning, many methods stack SAR and optical images along the channel dimension and learn a direct mapping to cloud-free targets using convolutional neural networks or generative adversarial networks. Although straightforward, this strategy ignores the physical discrepancy between SAR and optical imaging mechanisms. As a result, SAR speckles and thick-cloud artifacts may be indiscriminately propagated into the reconstruction pipeline. These pioneering methods have provided an important foundation for SAR-assisted cloud removal, but their reliance on direct concatenation makes them vulnerable to cross-modal feature entanglement, which can cause color deviation and limit texture recovery, especially under severe cloud occlusion~\cite{li2026deep}.

\begin{figure}[t]
    \centering
    \includegraphics[width=1\linewidth]{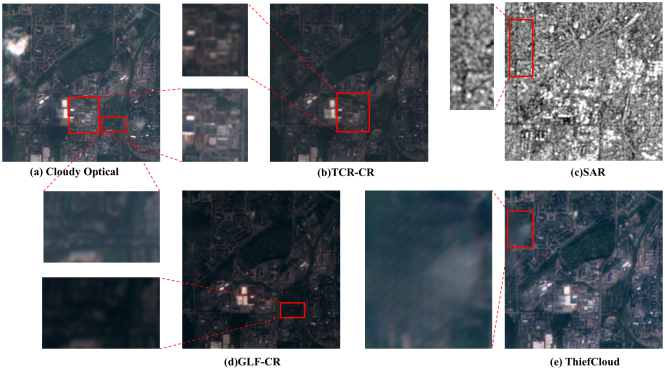}
    \caption{Visual illustration of the limitations of existing multimodal cloud removal methods. Although state-of-the-art methods such as (b) TCR-CR~\cite{wang2024tcrnet}, (d) GLF-CR~\cite{li2022glf}, and (e) ThiefCloud~\cite{li2025thiefcloud} can remove thick clouds from the (a) cloudy optical input, they still suffer from pixel degradation and over-smoothed textures. Due to speckle interference in the (c) SAR prior, existing networks struggle to reconstruct fine road structures and may produce striping artifacts and color deviation in the restored optical images.}
    \label{fig:intro_problem}
\end{figure}

The second group performs feature-level late fusion by modeling deeper spatial-spectral relationships between SAR and optical features. Representative strategies include attention-based fusion, conditional diffusion modeling, and two-flow spectral-spatial Transformer architectures~\cite{psfcr2026,wang2024iterative,yao2025adaptive}. For example, task-driven prompt learning introduces learnable degradation prompts to guide SAR-optical integration~\cite{tdpcr2026}. Despite their progress, these methods still suffer from representation bottlenecks in challenging scenarios. Since most of them lack explicit information-theoretic constraints, attention modules often learn implicit local responses rather than explicitly regulating noisy information flow. Moreover, when optimized mainly with pixel-level reconstruction losses, such as $L_1$ or $L_2$ losses, networks tend to produce averaged predictions, leading to over-smoothed textures or grid-like striping artifacts instead of sharp structures, as shown in Fig.~\ref{fig:intro_problem}. Diffusion-based methods and multi-temporal structure-aware diffusion models can alleviate blurring by modeling data distributions, but they often involve trade-offs among generation quality, sampling efficiency, and potential hallucinated content~\cite{zou2024diffcr,sader2026}. Uncertainty-aware approaches further suggest that reliable confidence estimation remains important for unconstrained generative reconstruction~\cite{ebel2023uncrtaints}.

To address these issues, we propose an Information Bottleneck-Driven High-Fidelity Network, termed IB-HFN, for SAR-assisted optical cloud removal. Our motivation is to decouple two conflicting objectives: suppressing heterogeneous SAR speckles and preserving high-frequency optical textures. Contrastive learning has shown effectiveness in compact feature learning for removing atmospheric degradation~\cite{wu2021contrastive}. Recent representation learning studies further demonstrate that combining contrastive learning with the variational information bottleneck (VIB) can reduce redundant correlations and noisy relevance~\cite{li2025contrastive,alemi2016deep}. Inspired by this principle, we tailor the information bottleneck mechanism to dense multimodal fusion. Instead of applying spatial VIB directly, which may disrupt two-dimensional spatial topology and introduce artifacts in image-to-image translation, we design a Spatial Information Bottleneck Fusion (SIBF) module. SIBF adopts an asymmetric routing strategy. Specifically, SAR features are compressed by a Channel-wise VIB (C-VIB) with a negative prior to suppress unstructured speckles, while a Local-Global Gate predicts clear-sky regions and routes reliable optical textures through a Dirac-initialized skip connection. This design separates SAR noise filtering from optical detail preservation. In addition, we introduce a joint optimization strategy that combines an internal feature-level compression penalty with an external image-level quadruple loss, jointly constraining reconstruction accuracy, structural similarity, spectral fidelity, and contrastive sharpness. A dynamic weight scheduling strategy coordinates these objectives to stabilize training and reduce hazy artifacts.

The main contributions of this paper are summarized as follows:
\begin{itemize}
    \item We propose IB-HFN, an information bottleneck-driven framework for SAR-assisted optical cloud removal. A dual-stream backbone preserves modality-specific representations until deep semantic fusion, reducing early cross-modal contamination. Selective Kernel fusion is further used in the decoder to aggregate multi-scale features.

    \item We design a Spatial Information Bottleneck Fusion module to address the conflict between SAR speckle suppression and optical texture preservation. SAR features are compressed by a Channel-wise VIB to filter unstructured speckles, while clear-sky optical details are adaptively routed through a Dirac-initialized skip connection guided by a Local-Global Gate.

    \item We introduce a joint optimization strategy with Dynamic Weight Scheduling. The objective integrates feature-level bottleneck regularization and an image-level quadruple loss, enabling stable compression, spectral preservation, structural consistency, and high-frequency texture recovery.

    \item Extensive experiments on the SEN12MS-CR dataset demonstrate that IB-HFN achieves strong robustness under spatio-temporal shifts. Compared with existing state-of-the-art methods, our approach better preserves deterministic SAR geometry and optical spectral fidelity.
\end{itemize}
\section{Related Work}
\label{sec:rw}

\subsection{Cloud Removal and Multimodal Fusion in Remote Sensing}

Optical remote sensing images are often degraded by clouds and shadows, which obscure land surface observations and limit downstream interpretation. To recover missing optical information, SAR assisted cloud removal has become an important direction because SAR can penetrate clouds and provide complementary geometric and structural priors. Early methods mainly rely on image concatenation or shallow spectral transformation~\cite{review2025cloud}, but they ignore the sensing gap between active microwave imaging and passive optical imaging, allowing SAR speckle noise to contaminate optical reconstruction. Recent deep models therefore shift to feature level fusion~\cite{zhou2025mmcanet}, where CNN or Transformer based spatial spectral attention dynamically selects multimodal cues~\cite{li2025transformer,igarss2025graph}. Representative methods exploit SAR edges~\cite{wang2024tcrnet,bui2024cloudaware}, efficient attention~\cite{huang2025attcr}, frequency prompting~\cite{huang2024dvpnet}, or physical atmospheric priors~\cite{li2025thiefcloud} to improve cloud region recovery.

Generative models further improve visual realism by modeling complex data distributions. Conditional diffusion has shown strong controllability in image and video generation tasks~\cite{shen2024advancing,shen2024imagpose,shenlong,shen2025boosting,shen2025imagedit,shen2025imagharmony,shen2025imagdressing,shen2025imaggarment}, and has been introduced into remote sensing cloud removal~\cite{zou2024diffcr,hu2025multimodal,chen2025sar}. However, diffusion models usually require iterative sampling and may hallucinate structures when optical information is severely missing. Although recent datasets support more systematic evaluation~\cite{chen2025cloud,cben2026}, existing fusion methods still lack explicit constraints on information flow. Attention based models may entangle SAR speckles with optical cloud residuals, while generative models may reduce structural reliability. This motivates a principled fusion mechanism that suppresses modality specific noise while preserving reliable optical textures.

\begin{figure*}[t]
    \centering
    \includegraphics[width=1\textwidth]{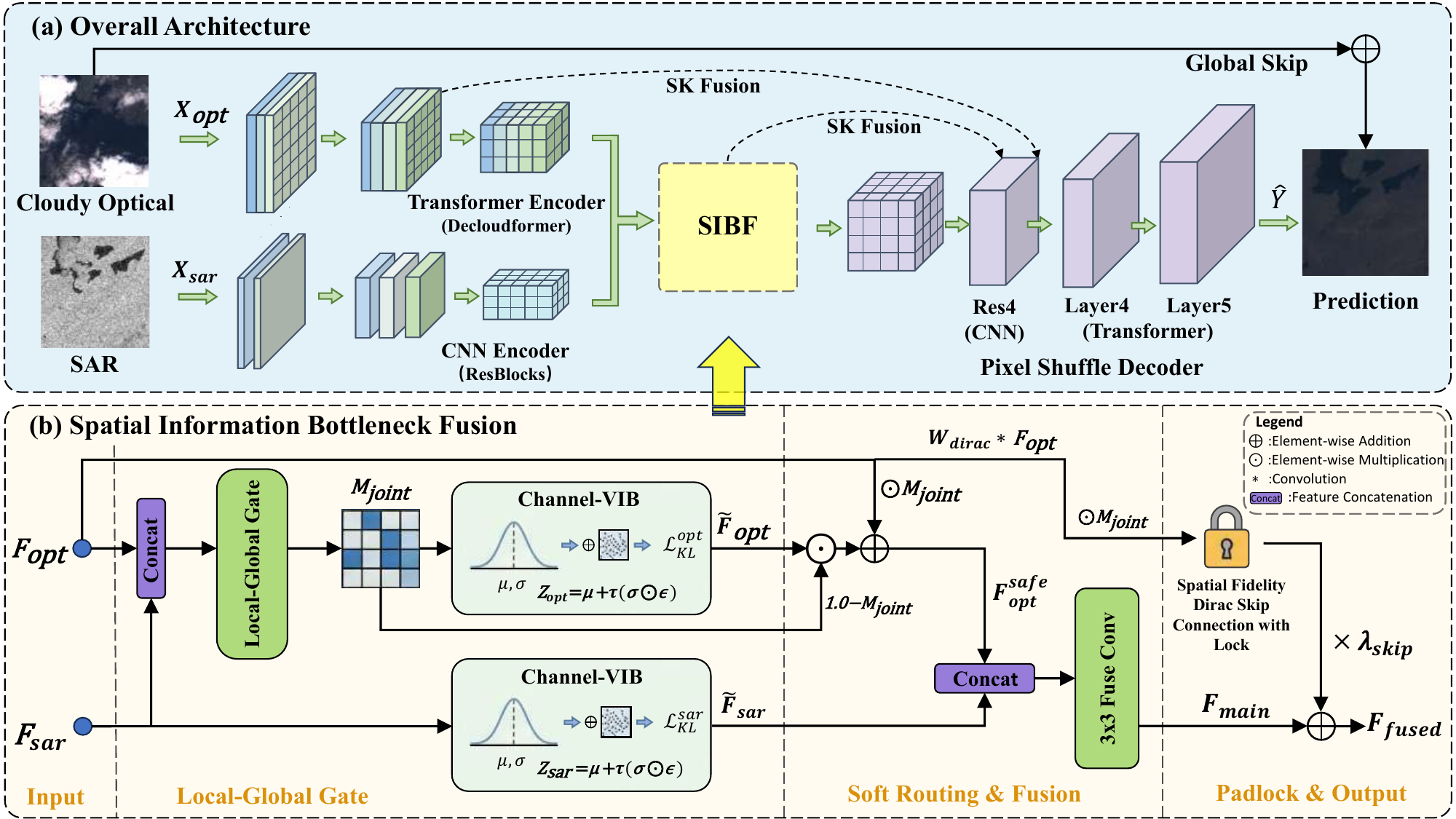}
    \caption{Overview of the proposed Information Bottleneck Driven High Fidelity Network. (a) IB-HFN adopts a dual stream architecture, where a Transformer based optical encoder and CNN based SAR encoder extract modality specific representations before deep fusion. Selective Kernel fusion is used in skip connections to aggregate multi scale features for the PixelShuffle decoder. (b) The Spatial Information Bottleneck Fusion module uses a Local Global Gate to predict the clear sky mask $M_{joint}$, applies Channel wise VIB to suppress modality specific noise through $\mathcal{L}_{KL}^{opt}$ and $\mathcal{L}_{KL}^{sar}$, and preserves high frequency optical details through a locked Dirac initialized skip connection.}
    \label{fig:architecture}
\end{figure*}

\subsection{Feature Decoupling and Information Bottleneck Theory}

Feature decoupling aims to separate task relevant cues from redundant or noisy factors. This is important for multimodal remote sensing, where SAR speckles, cloud residuals, seasonal variations, and spectral shifts are often entangled. Existing methods commonly rely on attention reweighting or feature alignment~\cite{wang2025mdfanet}, which improves feature selection but does not explicitly control the amount of information transmitted through the fusion path.

Information Bottleneck (IB) theory offers a principled way to balance representation compression and predictive sufficiency~\cite{tishby2015deep}. Recent studies show that bottleneck mechanisms can suppress task irrelevant noise and improve robustness~\cite{iclr2025projection}. IB based methods have been explored in vision transformer interpretation~\cite{coiba2025}, graph representation learning~\cite{ibcs2024}, remote sensing segmentation~\cite{shou2025graph}, and change detection~\cite{yin2025information}. Variational Information Bottleneck (VIB) further makes IB tractable by introducing stochastic latent variables~\cite{alemi2016deep}, and has been used for feature purification~\cite{ning2026refining} and hallucination mitigation in generative models~\cite{vibprobe2026}.

However, directly applying spatial VIB to dense cloud removal can damage two dimensional topology and suppress high frequency optical structures. We therefore shift the bottleneck from the spatial dimension to the channel dimension. The resulting Channel wise VIB filters noisy SAR responses while preserving spatial layout. Together with a clear sky gated Dirac skip connection, our SIBF module decouples SAR speckle suppression from optical texture preservation for high fidelity multimodal reconstruction.

\section{Proposed Methodology}\label{sec:method}

In this section, we elaborate on the proposed Information Bottleneck-Driven High-Fidelity Network (IB-HFN) for SAR-guided optical cloud removal. We first present the overall dual-stream architecture. Subsequently, to explicitly address the core dilemma of cross-modal feature entanglement—rigorously filtering heterogeneous SAR speckle noise while preventing information compression from destroying high-frequency clear-sky optical textures—we systematically restructure our methodology into three interconnected components: the Dual-Stream Backbone with SK Fusion to prevent early contamination, the Spatial Information Bottleneck Fusion (SIBF) module to physically isolate noise, and the Joint Optimization Strategy with Quadruple High-Fidelity Loss to navigate the perception-distortion tradeoff.

\subsection{Dual-Stream Backbone with SK Fusion}

Let $X_{opt} \in \mathbb{R}^{C_{opt} \times H \times W}$ and $X_{sar} \in \mathbb{R}^{C_{sar} \times H \times W}$ denote the cloudy optical multispectral image and the corresponding Synthetic Aperture Radar (SAR) image, respectively, where $C$ represents the channel dimension, and $H, W$ represent the spatial resolution. Our objective is to generate a high-fidelity, cloud-free optical image $\hat{Y} \in \mathbb{R}^{C_{opt} \times H \times W}$ that highly approximates the ground truth $Y$.

As illustrated in Fig. \ref{fig:architecture}(a), IB-HFN builds upon a decoupled dual-stream encoder-decoder baseline \cite{gu2025hpncr}. By maintaining independent optical features $F_{opt} \in \mathbb{R}^{D \times H' \times W'}$ through a Transformer-based encoder and SAR features $F_{sar} \in \mathbb{R}^{D \times H' \times W'}$ through CNN ResBlocks, it effectively prevents early modality contamination. More importantly, this strictly isolated feature extraction provides the structural prerequisite for our core asymmetric routing mechanism at the deep semantic stage.

Unlike the original baseline (HPN-CR) that rigidly concatenates $F_{opt}$ and $F_{sar}$ at the bottleneck—which inevitably injects SAR speckle noise into the optical domain—our IB-HFN completely restructures this core junction. However, universally applying information compression to filter this noise would inevitably smear high-frequency optical textures in genuine clear-sky regions. To break this dilemma, we substitute the baseline's naive concatenation with the proposed Spatial Information Bottleneck Fusion (SIBF) module. While both modalities enter this deep bottleneck, they are subjected to a fundamentally asymmetric routing topology. Specifically, SAR representations are unconditionally compressed to truncate unstructured speckle noise. Conversely, optical features are conditionally protected: while they participate in the bottleneck for semantic alignment, their pristine high-frequency details in clear-sky regions safely circumvent the severe compression via a spatial mask-guided direct connection.

To effectively bridge these safely bypassed hierarchical semantics with the decoding process, we incorporate Selective Kernel (SK) Fusion within the skip connections, which dynamically aggregates multi-scale features from varying receptive fields. Finally, a Pixel Shuffle Decoder (comprising a CNN-based Res4 block followed by cascaded Transformer-based Layer4 and Layer5 blocks) is employed to reconstruct the target cloud-free image.

\subsection{Spatial Information Bottleneck Fusion (SIBF) Module}

As detailed in Fig. \ref{fig:architecture}(b), the SIBF module is the core engine of IB-HFN for noise isolation and feature selection. It consists of a Channel-wise Variational Information Bottleneck (C-VIB), a Local-Global Gate, and a Dirac-initialized Skip Connection.

\subsubsection{Channel-wise Variational Information Bottleneck}
The Information Bottleneck (IB) principle asserts that an optimal representation should maximize mutual information with the target while minimizing mutual information with the input. Specifically, to prevent modality-specific noises from contaminating the fusion process, we symmetrically compress both $F_{opt}$ and $F_{sar}$ into stochastic latent variables $Z_{opt}$ and $Z_{sar}$. 

Taking the SAR branch as an example, to protect the 2D spatial geometry from fragmented degradation caused by independent channel-wise perturbations, we strictly confine the VIB process to the channel dimension. We first compute the channel-wise posterior mean $\mu_{sar}$ and log-variance $\log(\sigma_{sar}^2)$:
\begin{equation}
    [\mu_{sar}, \log(\sigma_{sar}^2)] = \mathcal{E}_{vib} \big( \operatorname{GAP}(F_{sar}) \big),
\end{equation}
where $\text{GAP}(\cdot)$ denotes Global Average Pooling, and $\mathcal{E}_{vib}$ denotes $1 \times 1$ convolutional layers. Using the reparameterization trick, the latent representation is sampled as:
\begin{equation}
    Z_{sar} = \mu_{sar} + \tau (\sigma_{sar} \odot \epsilon), \quad \epsilon \sim \mathcal{N}(0, I) ,
\end{equation}
where $\epsilon$ is random noise, $\tau$ is a hyperparameter scaling the noise variance to stabilize early training, and $\odot$ denotes element-wise multiplication with the channel attention broadcast across spatial dimensions. The sampled $Z_{sar}$ acts as a stochastic channel attention to modulate the original feature, yielding the purified feature $\tilde{F}_{sar} = F_{sar} \odot \sigma(Z_{sar})$, where $\sigma(\cdot)$ explicitly denotes the sigmoid activation function. Similarly, the optical branch utilizes an identical C-VIB architecture to generate its corresponding purified feature $\tilde{F}_{opt} = F_{opt} \odot \sigma(Z_{opt})$, alongside an independent KL divergence loss $\mathcal{L}_{KL}^{opt}$. This symmetric design ensures both modalities undergo consistent channel-wise noise purification. 

To enforce information compression, we introduce the Kullback-Leibler (KL) divergence loss, pulling the posterior distribution towards an isotropic Gaussian prior $\mathcal{N}(\mu_{prior}, I)$:
\begin{equation}
    \mathcal{L}_{KL}^{sar} = \frac{1}{2} \mathbb{E} \left[ (\mu_{sar} - \mu_{prior})^2 + \sigma_{sar}^2 - 1 - \log(\sigma_{sar}^2) \right].
\end{equation}
By setting a strictly negative prior ($\mu_{prior} < 0$), we shift the regularization anchor from an information-theoretic perspective. This forces the stochastic channels to default to a heavily suppressed state, ensuring the network is physically penalized for transferring excessive unstructured speckle noise, thereby retaining only the robust deterministic geometrical priors.

\subsubsection{Clear-Sky Masked Routing and Dirac Skip Connection}
While C-VIB excellently filters noise, applying compression universally causes high-frequency texture loss in cloud-free optical regions. To tackle this structural degradation, we dynamically predict a clear-sky mask $M_{joint} \in (0,1)^{H' \times W'}$ using a Local-Global Gate:
\begin{equation}
    M_{joint} = \text{Clamp} \Big( \sigma \big( \mathcal{F}_{gate}([F_{opt}, F_{sar}]) \big), \epsilon_{min}, \epsilon_{max} \Big),
\end{equation}
where $\mathcal{F}_{gate}$ aggregates local spatial features and global context, and we explicitly define clamping bounds $\epsilon_{min}$ and $\epsilon_{max}$ to stabilize the routing gradients. Specifically, values of $M_{joint}$ approaching the upper bound $\epsilon_{max}$ indicate genuine cloud-free regions, whereas values approaching the lower bound $\epsilon_{min}$ indicate heavily occluded areas. We leverage $M_{joint}$ to perform a soft routing, bypassing the VIB penalty for clear regions:
\begin{equation}
F_{opt}^{safe} = F_{opt} \odot M_{joint} + \tilde{F}_{opt} \odot (1 - M_{joint}).
\end{equation}
The main fusion is then executed as $F_{main} = \mathcal{F}_{fuse}([F_{opt}^{safe}, \tilde{F}_{sar}])$, where $\mathcal{F}_{fuse}$ denotes a spatial convolutional layer followed by Instance Normalization and LeakyReLU activation to safely bridge the domain gap.

To achieve sharpness in clear regions, we introduce a Clamped High-Frequency Skip Connection parameterized by a point-wise Dirac-initialized convolution $W_{dirac}$ (which mathematically sets the center weight to 1 and all others to 0, ensuring it initially acts as an exact identity mapping).  This connection acts as a pure identity mapping to structurally compensate the VIB module:
\begin{equation}
    F_{fused} = F_{main} + \lambda_{skip} \big( (W_{dirac} \ast F_{opt}) \odot M_{joint} \big),
\end{equation}
where $\lambda_{skip}$ is a learnable scale. The $M_{joint}$ lock ensures that pristine optical features are forcefully routed only in genuine clear-sky areas, perfectly decoupling noise filtration from texture preservation.

% =================== 插入图 3：联合重建约束 ===================
\begin{figure}[t]
    \centering
    \includegraphics[width=\columnwidth]{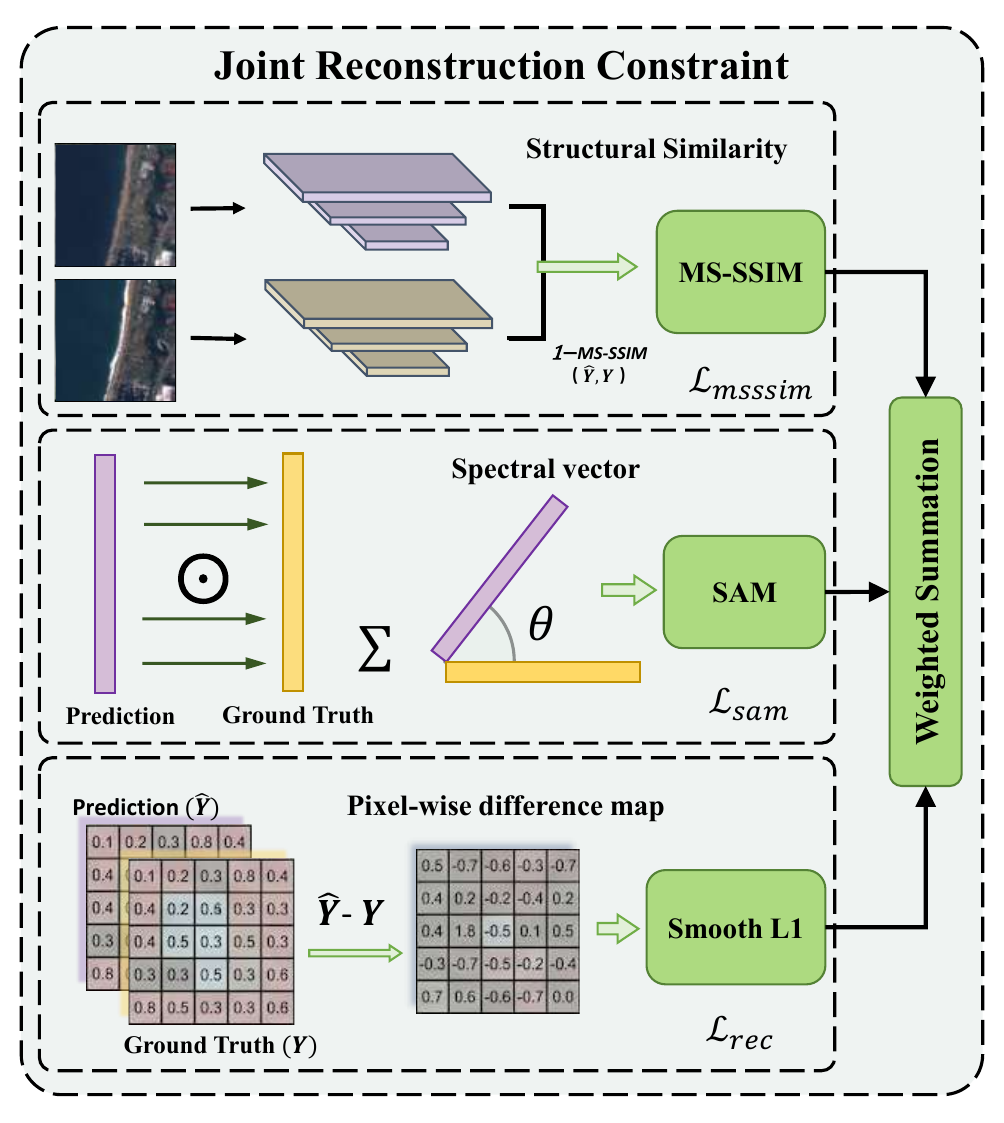}
    \caption{Illustration of the Joint Reconstruction Constraint. It integrates MS-SSIM ($\mathcal{L}_{msssim}$) using multi-scale image pyramids, Spectral Angle Mapper ($\mathcal{L}_{sam}$) via high-dimensional spectral vector alignment, and Smooth $L_1$ ($\mathcal{L}_{rec}$) based on pixel-wise difference maps ($\hat{Y} - Y$).}
    \label{fig:loss_joint}
\end{figure}
% ====================================================================

% =================== 插入图 4：多尺度对比学习 ===================
\begin{figure}[t]
    \centering
    \includegraphics[width=\columnwidth]{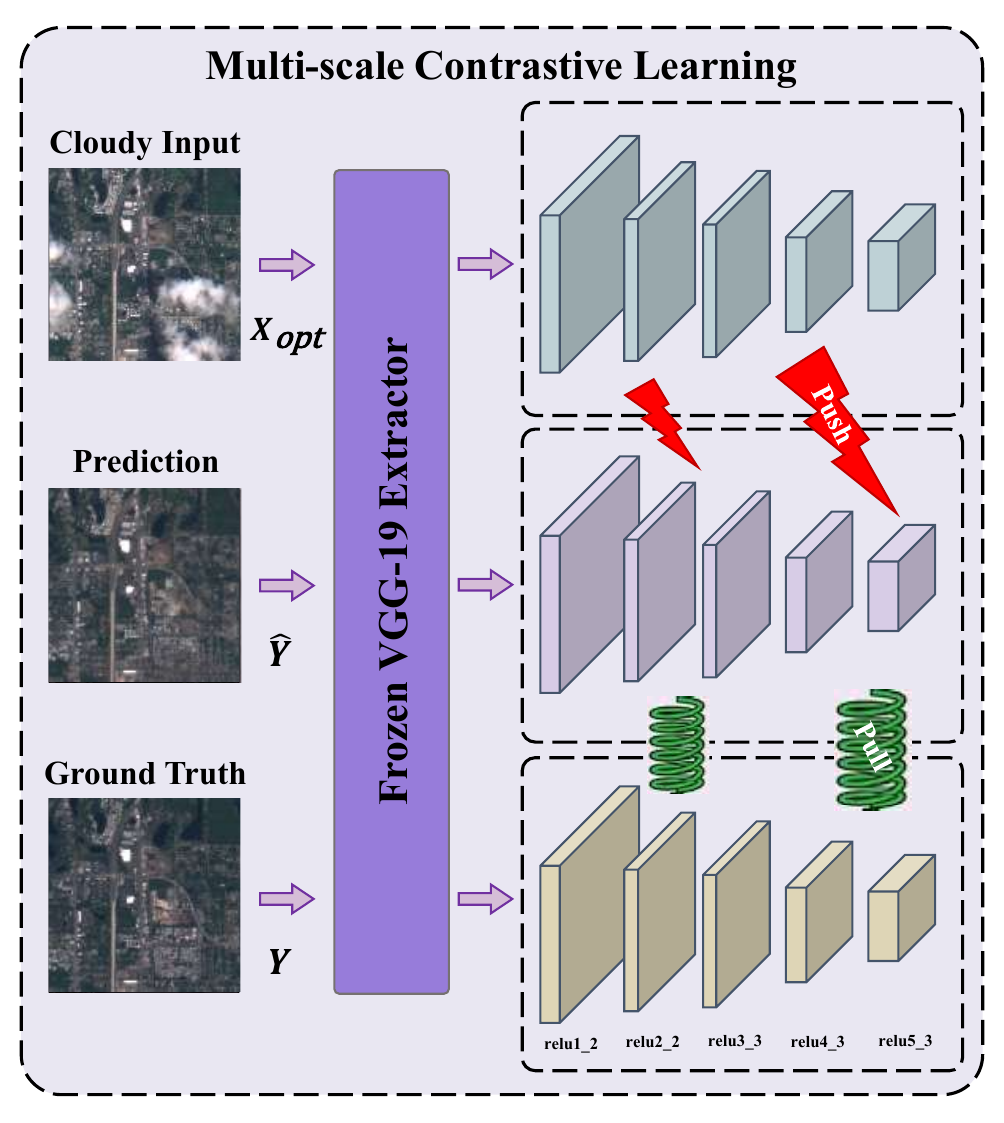}
    \caption{Illustration of the Multi-scale Contrastive Learning ($\mathcal{L}_{cr}$). It utilizes a shared frozen VGG-19 extractor to explicitly pull the prediction $\hat{Y}$ towards the clear ground truth $Y$ (Pull force) while pushing it away from the cloudy input $X_{opt}$ (Push force) across five hierarchical feature scales.}
    \label{fig:loss_contrastive}
\end{figure}
% ====================================================================
\subsection{Joint Optimization and Quadruple High-Fidelity Loss}

While the original baseline (HPN-CR) relies solely on a standard $L_1$ pixel-wise loss, this primitive constraint forces the network into a ``mean-value trap,'' causing the restored images to appear smoothed and spectrally distorted. Furthermore, optimizing spatial reconstruction simultaneously with our newly introduced deep information compression can lead to severe gradient conflicts. To fundamentally resolve these baseline limitations, we completely upgrade the optimization paradigm into a joint strategy that explicitly divides the five optimization constraints into two distinct levels: an internal feature-level compression penalty ($\mathcal{L}_{KL}$) and an external Image-level Quadruple Loss.
\subsubsection{Image-level Quadruple Loss}
As illustrated in Fig. \ref{fig:loss_joint} and Fig. \ref{fig:loss_contrastive}, the Image-level Quadruple Loss encompasses spatial, spectral, structural, and semantic dimensions to guarantee the fidelity of the perception-level output.

First, to prevent severe outliers and ensure stable gradient backpropagation, we employ the Smooth $L_1$ loss ($\mathcal{L}_{rec}$) as our primary spatial reconstruction constraint:
\begin{equation}
\resizebox{0.91\columnwidth}{!}{$
    \mathcal{L}_{rec} = \frac{1}{N C_{opt}} \sum_{i=1}^{N} \sum_{c=1}^{C_{opt}} 
    \begin{cases} 
    \frac{1}{2} (\hat{Y}_{i,c} - Y_{i,c})^2, & \text{if } |\hat{Y}_{i,c} - Y_{i,c}| < 1 \\ 
    |\hat{Y}_{i,c} - Y_{i,c}| - \frac{1}{2}, & \text{otherwise} 
    \end{cases},
$}
\end{equation}
where $N = H \times W$ is the total number of spatial pixels, $C_{opt}$ is the number of spectral channels, and $\hat{Y}_{i,c}$ and $Y_{i,c}$ denote the pixel values at the $c$-th channel of the $i$-th spatial location.

Second, to explicitly eliminate the spectral distortion caused by heterogeneous SAR fusion, we introduce the Spectral Angle Mapper (SAM) loss, which strictly constrains the angle between the predicted and target spectral vectors:
\begin{equation}
    \mathcal{L}_{sam} = \frac{1}{N} \sum_{i=1}^{N} \arccos \left( \frac{\hat{Y}_i \cdot Y_i}{\|\hat{Y}_i\|_2 \|Y_i\|_2 + \xi} \right),
\end{equation}
where $N = H \times W$ is the total number of spatial pixels, $\hat{Y}_i, Y_i \in \mathbb{R}^{C_{opt}}$ represent the spectral vectors at the $i$-th pixel location, and $\xi$ is a tiny stabilizer. 

Third, to suppress residual grid artifacts, we deploy the Multi-Scale Structural Similarity (MS-SSIM) loss:
\begin{equation}
    \mathcal{L}_{msssim} = 1 - \text{MS-SSIM}(\hat{Y}, Y).
\end{equation}

Fourth, to cure the smoothing effect in clear regions, we migrate Contrastive Learning into the restoration domain. We establish a high-dimensional feature space using a pre-trained VGG19 network \cite{wu2021contrastive}. Treating $\hat{Y}$ as anchor, the clear target $Y$ as positive sample, and the cloudy input $X_{opt}$ as negative sample, contrastive loss $\mathcal{L}_{cr}$ is formulated as:
\begin{equation}
    \mathcal{L}_{cr} = \sum_{j=1}^{K} \omega_j \frac{\| \Phi_j(\hat{Y}) - \Phi_j(Y) \|_1}{\| \Phi_j(\hat{Y}) - \Phi_j(X_{opt}) \|_1 + \eta},
\end{equation}
where $K=5$ corresponds to the extracted VGG19 features (relu1\_2, relu2\_2, relu3\_3, relu4\_3, relu5\_3), and $\eta$ prevents division by zero. To prioritize deep semantic alignment over shallow pixel differences, the scale weights $\omega_j$ are deliberately set to exponentially increasing values: $[1/32, 1/16, 1/8, 1/4, 1.0]$.This regularization acts as a complementary high-frequency sharpener: it pulls the prediction towards the clear texture distribution while pushing it away from the hazy distribution.

\subsubsection{Dynamic Weight Scheduling (DWS)}
The overall optimization objective tightly couples the external image-level Quadruple Loss with the internal feature-level SIBF compression penalty:
\begin{equation}
    \mathcal{L}_{total} = (\mathcal{L}_{rec} + \mathcal{L}_{msssim} + \mathcal{L}_{sam}) + \alpha \mathcal{L}_{cr} + \beta(t) (\mathcal{L}_{KL}^{sar} + \mathcal{L}_{KL}^{opt}),
\end{equation}
where $\alpha$ is a static trade-off parameter for the high-frequency sharpener, and $\beta(t)$ is an epoch-dependent dynamic weight for the joint VIB constraint.

To prevent gradient conflicts, we employ a Dynamic Weight Scheduling (DWS) strategy. During the initial warm-up epochs, $\beta(t)$ is set to 0, allowing the network to construct a stable cross-modal spatial mapping without the burden of information compression. Subsequently, utilizing cosine annealing, $\beta(t)$ is smoothly increased to physically truncate unstructured SAR noise, while the static $\alpha$ continuously provides gradient momentum to force the reconstruction of sharp edges. This dynamic game successfully achieves the ultimate balance between deep spatial decoupling and noise forgetting \cite{alemi2016deep}.

\section{Experiments and Analysis}\label{sec:experiments}

To comprehensively evaluate the effectiveness of the proposed Information Bottleneck-Driven High-Fidelity Network (IB-HFN), extensive quantitative and qualitative experiments are conducted. This section details the dataset preparation, evaluation metrics, and implementation settings. Subsequently, we compare our method with state-of-the-art (SOTA) multimodal cloud removal algorithms. Finally, rigorous ablation studies are provided that progressively validate dual-stream backbone design, structural decoupling of the SIBF module, and the optimization dynamics of the Joint Optimization Strategy to explicitly interpret the physical mechanism of the proposed framework.
% =================== table1数据集划分表格 ===================
\begin{table}[t]
\centering
\caption{Temporal-Semantic Shift Splitting on the SEN12MS-CR Dataset. Note: Rather than employing strict spatial isolation, this protocol defines (Scene, Season) tuples as independent samples, deliberately constructing an extreme stress test against spatial memorization.}
\label{tab:dataset_split}
\renewcommand{\arraystretch}{1.5} % 行高稍微再加大一点，配合变大的主字体
\resizebox{\columnwidth}{!}{
\begin{tabular}{c c c c >{\centering\arraybackslash}m{3.2cm}}
\toprule
\textbf{Subset} & \textbf{Season} & \textbf{Scenes} & \textbf{Patches (\%)} & \textbf{Scene IDs} \\
\midrule
\multirow{9}{*}{\textbf{Train}} 
 & Spr. & 30 & 21,789 (22.0) & \tiny 6, 8, 15, 17, 21, 31, 40, 44-45, 58, 63, 75, 77, 100-101, 106, 109, 117, 119-121, 123-124, 126, 128, 132, 134, 140-142 \\
 & Sum. & 39 & 26,933 (27.2) & \tiny 4, 7, 11, 15, 17, 19, 25, 36, 40, 42-43, 47, 55-56, 72-73, 76, 80, 86-87, 93, 95, 100-102, 113-115, 119, 121, 123-125, 133, 137, 139, 143, 146-147 \\
 & Fall & 52 & 37,310 (37.7) & \tiny 1, 3-4, 6, 11, 14, 19, 22, 27-28, 31, 33, 35, 37, 39-42, 64-65, 71, 77, 81-83, 85, 88, 91, 93, 100, 104-105, 107, 109-110, 112, 114, 116, 119-120, 122, 125, 133-136, 141-142, 144, 147-149 \\
 & Win. & 19 & 13,004 (13.1) & \tiny 8, 21-22, 42, 47, 49, 59, 61, 63-64, 68, 75, 84, 104, 107, 112, 116, 135, 146 \\
\cmidrule{2-5}
 & \textit{Subtotal} & \textit{140} & \textit{99,036} & - \\
\midrule
\multirow{5}{*}{\textbf{Val}}
 & Spr. & 3 & 1,606 (14.8) & \tiny 1, 9, 97 \\
 & Sum. & 5 & 3,079 (28.5) & \tiny 31, 89-90, 120, 127 \\
 & Fall & 3 & 1,926 (17.8) & \tiny 30, 57, 128 \\
 & Win. & 6 & 4,204 (38.9) & \tiny 25, 62, 81, 94, 102, 130 \\
\cmidrule{2-5}
 & \textit{Subtotal} & \textit{17} & \textit{10,815} & - \\
\midrule
\multirow{5}{*}{\textbf{Test}}
 & Spr. & 8 & 5,722 (46.3) & \tiny 26, 39, 66, 110, 113, 115, 145, 147 \\
 & Sum. & 5 & 3,815 (30.8) & \tiny 27, 126, 132, 135, 140 \\
 & Fall & 3 & 1,705 (13.8) & \tiny 26, 131, 139 \\
 & Win. & 2 & 1,125 (9.1)  & \tiny 55, 108 \\
\cmidrule{2-5}
 & \textit{Subtotal} & \textit{18} & \textit{12,367} & - \\
\midrule
\textbf{Global} & \textit{Total} & \textbf{175} & \textbf{122,218} & - \\
\bottomrule
\end{tabular}
}
\end{table}
% ============================================================

\subsection{Dataset Preparation and Evaluation Metrics}
All experiments are conducted on the widely recognized multimodal cloud removal dataset, SEN12MS-CR \cite{2022sen12ms}. This large-scale dataset provides strictly co-registered image triplets: cloudy optical images (Sentinel-2), corresponding SAR images (Sentinel-1), and cloud-free target optical images. 

Temporal-Semantic Shift Splitting and Extreme Stress Test: In standard remote sensing cloud removal tasks, most existing methods adopt a naive random patch-level splitting strategy. However, this conventional paradigm suffers from severe spatial data leakage, where adjacent overlapping patches from the same large-scale image inevitably leak into both the training and testing sets, leading to artificially inflated performance metrics. To overcome this critical flaw and construct a genuinely rigorous benchmark, we completely abandon the random shuffle approach and adopt a Scene-Level Temporal-Semantic Shift Splitting strategy based on (Scene, Season) tuples. First, our protocol strictly operates at the macroscopic scene level, explicitly preventing the trivial patch-level spatial leakage common in previous works. Second, as detailed in Table \ref{tab:dataset_split}, rather than employing a conventional strict spatial isolation, we deliberately distribute the same geographical scene acquired in different seasons across the training and testing sets (e.g., Scene 147 appears in Train-Summer but is evaluated in Test-Spring). This deliberate design upgrades the evaluation from a simple spatial interpolation task to an extreme temporal-semantic stress test. It forces the network to confront significant temporal semantic shifts (e.g., memorizing a snow-free winter layout during training but being forced to reconstruct its snow-covered target during testing). By explicitly penalizing networks that take shortcuts via static spatial memorization, this rigorous protocol ensures that the reported performance genuinely reflects the SIBF module's capability to extract deterministic SAR geometric priors rather than hallucinating memorized seasonal textures.

To objectively evaluate reconstruction quality, we employ four widely used full-reference metrics: Peak Signal-to-Noise Ratio (PSNR), Structural Similarity Index Measure (SSIM), Spectral Angle Mapper (SAM), and Mean Absolute Error (MAE). While PSNR and MAE measure absolute pixel-level fidelity, SSIM evaluates the preservation of high-frequency geometric structures. Crucially, SAM assesses color and spectral fidelity; a lower SAM explicitly indicates less modal interference (e.g., optical color distortion caused by SAR speckle contamination), which is a primary optimization objective of our decoupled architecture.

\subsection{Implementation Details and Optimization Strategy}
The proposed IB-HFN is implemented in PyTorch and trained on an NVIDIA RTX 4090 GPU. Specifically, the network employs a strictly decoupled dual-stream backbone: a Window-Attention-based Transformer (Decloudformer) is utilized to extract optical features, while robust CNN ResBlocks are employed for SAR features. The spatial resolution of the input image patches is cropped to $256 \times 256$. We optimize the network using the AdamW optimizer with a batch size of 24. To maximize computational efficiency and optimize memory footprint, Automatic Mixed Precision (AMP) is employed during training. For the core SIBF module, the hyperparameters detailed in Section \ref{sec:method} are empirically configured as follows: the noise scaling factor is set to $\tau=0.1$, the negative prior is initialized at $\mu_{prior}=-2.0$, and the clamping bounds for the clear-sky mask are defined as $\epsilon_{min}=0.05$ and $\epsilon_{max}=0.95$. All reported quantitative metrics represent the average of three independent training runs with different random seeds (expressed as mean $\pm$ std).

Joint Optimization Implementation (DWS and Cosine Annealing): Recognizing the severe perception-distortion tradeoff in dense fusion tasks, we implement the Dynamic Weight Scheduling (DWS) strategy. We utilize a Cosine Annealing Learning Rate Scheduler, smoothly decaying from $1 \times 10^{-4}$ to $1 \times 10^{-6}$ over 50 epochs. For the joint loss constraints, the contrastive weight $\alpha$ is kept static at $0.05$ throughout the training process to stably carve high-frequency textures without overpowering the primary pixel-level reconstruction. Meanwhile, we orchestrate the compression penalty $\beta(t)$ through a precise three-stage game: 
(1) Warm-up Stage (Epochs 1-15): The VIB KL penalty $\beta(t)$ is set to 0. The network relies purely on the base reconstruction constraints to establish a solid cross-modal spatial alignment. 
(2) Annealing Stage (Epochs 16-35): $\beta(t)$ is smoothly annealed to a maximum of 0.001. This gradual penalty forces the network to start squeezing out heterogeneous SAR speckles while preserving the carved geometric edges. 
(3) Purification Stage (Epochs 36-50): $\beta(t)$ is maintained at its maximum penalty of 0.001. This ensures a persistent physical truncation of unstructured noise during the final fine-tuning, successfully achieving the ultimate perception-distortion balance.

% =================== 插入图 4：SOTA 视觉对比图 ===================
\begin{figure*}[t]
    \centering
    \includegraphics[width=1\textwidth]{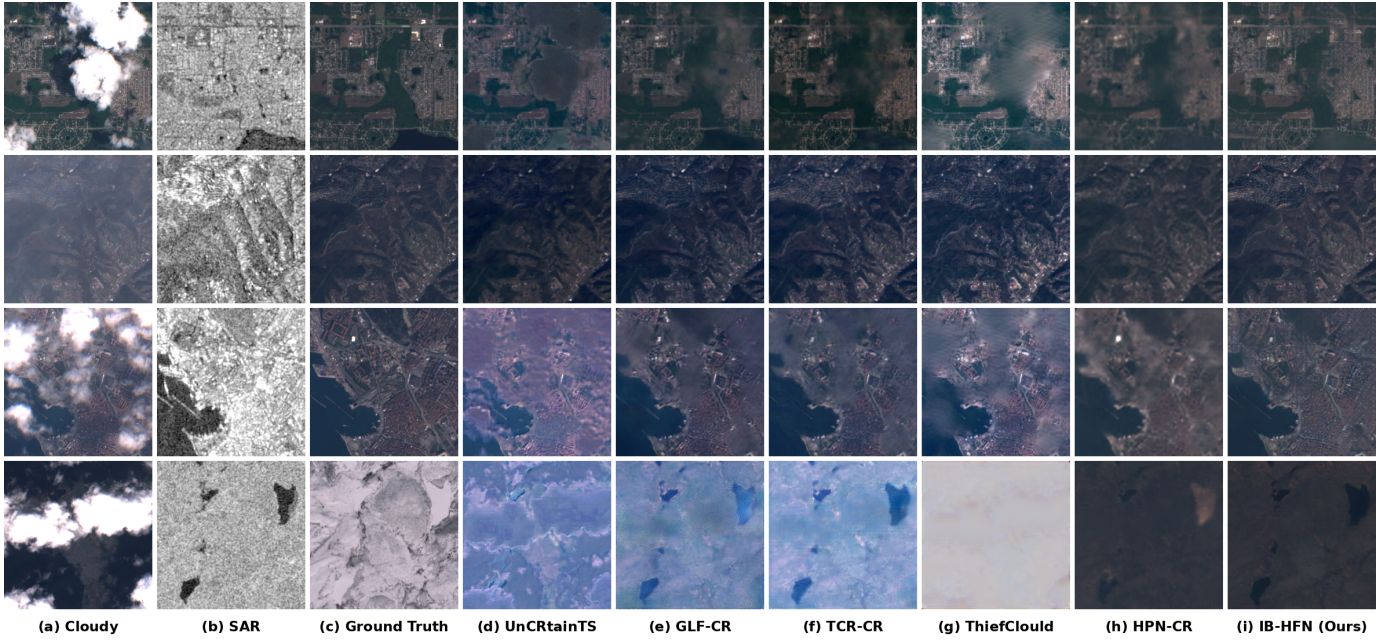} % SOTA大图文件
    \caption{Qualitative comparison under extreme thick-cloud scenarios with temporal semantic shifts. Recent state-of-the-art methods, such as UnCRtainTS (d) and GLF-CR (e), suffer from SAR speckle contamination (evident in blurry, noisy edges). Notably, in cases where the Ground Truth (c) exhibits snow cover absent in the cloudy inputs, methods like GLF-CR and TCRNet (f) tend to artificially shift their color distributions to match the target. In contrast, our IB-HFN (i) and HPN-CR (h) strictly adhere to the deterministic SAR physical priors to reconstruct the authentic non-snow landscape. Consequently, IB-HFN recovers the occluded complex topologies (e.g., coastlines, water bodies, and urban layouts) without hallucinating seasonal features, exhibiting clear edges and pure, physically accurate color distributions.}
    \label{fig:sota_visual}
\end{figure*}
% =================================================================

% --- TABLE 2: SOTA Comparison ---
\begin{table}[t]
\centering
\caption{Quantitative Comparison with SOTA Methods on SEN12MS-CR under Scene-Level Extreme Splitting. Best results are shown in \textbf{bold}.}
\label{tab:sota_compare}
\renewcommand{\arraystretch}{1.3} % 增加行高，提升呼吸感
\resizebox{\columnwidth}{!}{
\begin{tabular}{c c c c c}
\toprule
\textbf{Method} & \textbf{PSNR $\uparrow$} & \textbf{SSIM $\uparrow$} & \textbf{SAM $\downarrow$} & \textbf{MAE $\downarrow$} \\
\midrule
Cloudy  & 17.56 & 0.6687 & 12.43 & 0.1594 \\
UnCRtainTS \cite{ebel2023uncrtaints} & 27.40 & 0.8263 & 8.95 & 0.0511 \\
GLF-CR \cite{li2022glf}              & 28.29 & 0.8588 & 8.34 & 0.0492 \\
TCRNet \cite{wang2024tcrnet}         & 28.15 & 0.8689 & 7.73 & 0.0468 \\
ThiefCloud \cite{li2025thiefcloud}   & 24.95 & 0.7792 & 9.04 & 0.0690 \\
HPN-CR \cite{gu2025hpncr}            & 28.26 & 0.8509 & 7.88 & 0.0484 \\ 
\midrule
\textbf{IB-HFN (Ours)} & \textbf{29.12}$_{\pm 0.09}$ & \textbf{0.8794}$_{\pm 0.004}$ & \textbf{7.15}$_{\pm 0.06}$ & \textbf{0.0458}$_{\pm 0.003}$ \\
\bottomrule
\end{tabular}
}
\end{table}
%----------------------------
\subsection{Comparison with State-of-the-Art Methods}

To demonstrate the superiority of the proposed framework, we compare it against representative SOTA methods, including spatial-temporal sequence models (e.g., UnCRtainTS), multi-scale attention networks (e.g., GLF-CR, TCRNet \cite{wang2024tcrnet}), prior-driven approaches (e.g., ThiefCloud \cite{li2025thiefcloud}), and the unified spatial baseline network (HPN-CR \cite{gu2025hpncr}). 

Rationale for Baseline Selection: In this study, we deliberately select HPN-CR as our foundational baseline. Unlike methods that prematurely entangle SAR and optical features through dense early-stage cross-attention, HPN-CR maintains a strictly decoupled, dual-stream parallel topology until the deep semantic stage. This clean, late-fusion architecture provides a pristine testing ground that perfectly accommodates our SIBF module. By injecting the C-VIB layer precisely at the deep bottleneck, we explicitly filter out modality-specific noises before any spatial cross-contamination occurs. All comparative methods are retrained under our rigorous spatio-temporal splitting protocol.

\subsubsection{Quantitative Evaluation}
The quantitative results are reported in Table \ref{tab:sota_compare}. Due to the extreme temporal semantic shifts of our stress-test set, the absolute metrics for all baseline methods experience a global degradation. Early fusion networks perform suboptimally, particularly in SAM, due to speckle entanglement. While recent attention-based methods improve PSNR, they still suffer from spectral distortion. However, under such harsh conditions, IB-HFN achieves the best performance across all metrics (PSNR: 29.12 dB, SAM: 7.15). The significantly widened performance gap under this severe occlusion setting demonstrates its capability in decoupling modality noise.

\subsubsection{Qualitative Visual Comparison}
As visualized in Fig. \ref{fig:sota_visual}, the original baseline HPN-CR (h) suffers from modality collapse under thick clouds. While recent SOTA methods like GLF-CR (e) and TCRNet (f) manage to enhance the overall brightness, they introduce spectral shifts and false color artifacts (e.g., unnatural purplish tints in water bodies). In contrast, our IB-HFN (i) effectively restores the degraded baseline. Guided by the Quadruple Loss and the C-VIB physical bottleneck, IB-HFN filters out the SAR speckle interference while selectively routing optical priors, exhibiting fine sharpness and pure color distributions.

A Note on Temporal Semantic Shifts: It is worth noting that a visual disparity occasionally appears between the prediction and the Ground Truth (e.g., the target GT captures a snow-covered winter scene, while the input images were acquired during a snow-free winter period). Even under such temporal misalignment, our IB-HFN reconstructs the genuine snow-free landscape based on the deterministic physical backscattering of SAR, rather than hallucinating snow to artificially match the GT. This explicitly proves the model's loyalty to structural priors over spatial memorization.

\subsection{Comprehensive Ablation Studies}
We evaluate how our three interconnected components progressively contribute to the overall system, followed by fine-grained micro-ablations to deconstruct the internal physical mechanisms.

% --- TABLE 3: SIBF Structural Ablation ---
\begin{table}[t]
\centering
\caption{Ablation on SIBF Structural Components.}
\label{tab:sibf_ablation}
\renewcommand{\arraystretch}{1.3} 
\begin{tabular}{l c c c}
\toprule
\textbf{Architectural Evolution} & \textbf{PSNR $\uparrow$} & \textbf{SSIM $\uparrow$} & \textbf{SAM $\downarrow$} \\
\midrule
Baseline & 28.26 & 0.8509 & 7.88 \\
+ w/o C-VIB & 28.38 & 0.8661 & 7.29 \\
+ w/o Dirac Skip Connection & 28.47 & 0.8649 & 7.42 \\
\textbf{IB-HFN (Ours)} & \textbf{29.12} & \textbf{0.8794} & \textbf{7.15} \\
\bottomrule
\end{tabular}
\end{table}
%==================================================

\subsubsection{Ablation on SIBF Structural Components}
\begin{figure}[t!]
    \centering
    \includegraphics[width=1\linewidth]{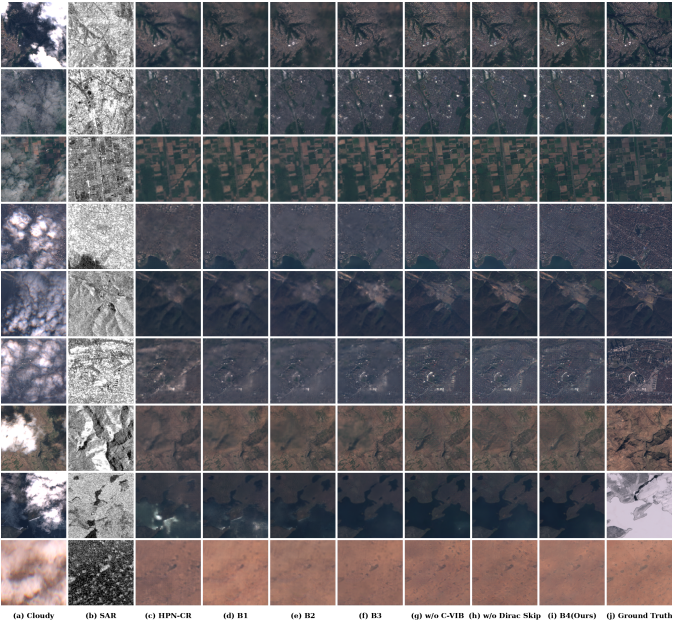} % 消融图
     \vspace{-0.3cm}
    \caption{Visual ablation of the Quadruple High-Fidelity Loss and SIBF structural components. Compared to the baseline HPN-CR (c), relying solely on the $L_1$ loss (Variant B1, d) still traps the network in a mean-value smoothing effect. The progressive integration of SAM (Variant B2, e) corrects spectral color shifts, while MS-SSIM (Variant B3, f) enhances structural alignment. Furthermore, ablating the structural modules reveals that removing the C-VIB layer (g) leads to SAR speckle leakage and color distortion, whereas discarding the Dirac Skip connection (h) inevitably degrades high-frequency clear-sky textures. By synergizing all optimization and structural components, our full IB-HFN (Variant B4, i) effectively mitigates hazy artifacts, purifies heterogeneous noise, and restores intricate geometric details that closely align with the Ground Truth (j).}
    \label{fig:loss_ablation}
\end{figure}

We first validate the architectural necessity of the SIBF module. As reported in Table \ref{tab:sibf_ablation}, the baseline achieves a PSNR of 28.26 dB but suffers from severe modal entanglement (SAM: 7.88). To integrate features, if we simply omit the physical compression layer (w/o C-VIB)---which essentially degenerates the fusion mechanism into a simple direct spatial concatenation---the network lacks an effective physical noise filter. Consequently, SAR speckle noise inevitably leaks into the optical domain, leading to a suboptimal SAM of 7.29 and visible color distortion as shown in Fig. \ref{fig:loss_ablation}(g). Conversely, when we apply information compression but forcefully remove the physical bypass (w/o Dirac Skip Connection), the mandatory bottleneck severely smooths the uncorrupted high-frequency clear-sky textures (Fig. \ref{fig:loss_ablation}(h)). Ultimately, our full IB-HFN achieves the best performance, validating that the synergistic combination of C-VIB (for denoising) and the Dirac Skip Connection (for detail preservation) is structurally indispensable.
% =================== 插入图 6：Loss 视觉消融演进图 ===================
% ====================================================================

% --- TABLE 4: Micro Ablation of SIBF Hyperparameters ---
\begin{table}[t]
\centering
\caption{Micro Ablation I: Hyperparameter Dynamics of C-VIB.}
\label{tab:sibf_micro}
\renewcommand{\arraystretch}{1.15} 
\begin{tabular}{c cc ccc}
\toprule
\textbf{Variant} & $\boldsymbol{\mu_{prior}}$ & $\boldsymbol{\beta_{max}}$ & \textbf{PSNR $\uparrow$} & \textbf{SSIM $\uparrow$} & \textbf{SAM $\downarrow$} \\
\midrule
1 & $0.0$  & $10^{-3}$ & 28.84 & 0.8690 & 7.41 \\
2 & $-1.0$ & $10^{-3}$ & 28.98 & 0.8730 & 7.28 \\
3 & $-2.0$ & $10^{-2}$ & 28.15 & 0.8214 & 8.52 \\
4 & $-2.0$ & $10^{-4}$ & 28.52 & 0.8510 & 7.95 \\
\midrule
\textbf{Ours} & $\mathbf{-2.0}$ & $\mathbf{10^{-3}}$ & \textbf{29.12} & \textbf{0.8794} & \textbf{7.15} \\
\bottomrule
\end{tabular}
\end{table}
%============================================

\subsubsection{Micro Ablation I: Hyperparameter Dynamics of C-VIB}
As detailed in Table \ref{tab:sibf_micro}, the SIBF module relies on a delicate parameter balance within C-VIB to achieve stochastic channel compression. Setting a strongly negative prior mean ($\mu_{prior} = -2.0$) forces the stochastic channels to default to a suppressed state, acting as a strict physical threshold to eradicate modal entanglement. If we use a standard prior ($\mu_{prior} = 0.0$, Variant 1) or an intermediate prior ($-1.0$, Variant 2), the network lacks this sufficient suppression mechanism, allowing SAR speckle noise to leak through and leading to suboptimal spectral fidelity (e.g., SAM worsens to 7.41 in Variant 1).

Regarding the information compression penalty ($\beta_{max}$), an excessively aggressive bottleneck leads to over-compression ($\beta_{max} = 10^{-2}$, Variant 3) of the deterministic backbone, causing a severe drop in PSNR (28.15 dB). Conversely, setting a weak constraint results in under-compression ($\beta_{max} = 10^{-4}$, Variant 4), which fails to adequately penalize the high-entropy SAR granular noise, leading to poor structural and spectral recovery. The optimal balance is achieved by our final model ($\mu_{prior} = -2.0$ and $\beta_{max} = 10^{-3}$), yielding the highest overall quantitative performance.
\subsubsection{Micro Ablation II: Image-level Quadruple Loss Mechanism}
To isolate the exact contribution of each optimization objective, we progressively add loss terms, as reported in Table \ref{tab:loss_micro} and visually corroborated in Fig. \ref{fig:loss_ablation}. Starting from Smooth $L_1$ (Variant B1, see Fig. \ref{fig:loss_ablation}(d)), the network still suffers from over-smoothed textures. Incorporating SAM (Variant B2) explicitly corrects spectral color shifts, and MS-SSIM (Variant B3) improves structural alignment. Finally, deploying the Multi-Scale Contrastive Loss (Variant B4) serves as the high-frequency sharpener, yielding the best perception-distortion balance and restoring crisp geometric details.
% --- TABLE 5: Loss Micro Ablation ---
\begin{table}[t]
\centering
\caption{Micro Ablation II: Deconstruction of the Quadruple Loss Mechanism.}
\label{tab:loss_micro}
\renewcommand{\arraystretch}{1.3} 
\resizebox{\columnwidth}{!}{
\begin{tabular}{c cccc ccc}
\toprule
\textbf{Model} & $\boldsymbol{L_{rec}}$ & $\boldsymbol{L_{sam}}$ & $\boldsymbol{L_{msssim}}$ & $\boldsymbol{L_{cr}}$ & \textbf{PSNR $\uparrow$} & \textbf{SSIM $\uparrow$} & \textbf{SAM $\downarrow$} \\
\midrule
B1 & \ding{51} & \ding{55} & \ding{55} & \ding{55} & 28.76 & 0.8651 & 7.60 \\
B2 & \ding{51} & \ding{51} & \ding{55} & \ding{55} & 28.89 & 0.8653 & 7.59 \\
B3 & \ding{51} & \ding{51} & \ding{51} & \ding{55} & 29.01 & 0.8675 & 7.56 \\
\textbf{B4 (Ours)} & \ding{51} & \ding{51} & \ding{51} & \ding{51} & \textbf{29.12} & \textbf{0.8794} & \textbf{7.15} \\
\bottomrule
\end{tabular}
}
\end{table}
%===================================================
% --- TABLE 6: DWS Micro Ablation ---
\begin{table}[t]
\centering
\caption{Micro Ablation III: Dynamic Weight Scheduling (DWS) and Annealing.}
\label{tab:dws_micro}
\renewcommand{\arraystretch}{1.3} % 保持呼吸感行高
\resizebox{\columnwidth}{!}{
\begin{tabular}{c c c c c}
\toprule
\textbf{Scheduling Configuration} & $\boldsymbol{T_{warmup}}$ & \textbf{PSNR $\uparrow$} & \textbf{SSIM $\uparrow$} & \textbf{SAM $\downarrow$} \\
\midrule
Hard Truncation ($\beta(t)$ Step Jump)   & -- & 28.61 & 0.8652 & 7.63 \\
Early Smooth Annealing                   & 0  & 28.58 & 0.8605 & 7.71 \\
Late Smooth Annealing                    & 30 & 28.85 & 0.8715 & 7.45 \\
\textbf{Proposed DWS (Ours)}             & \textbf{15} & \textbf{29.12} & \textbf{0.8794} & \textbf{7.15} \\
\bottomrule
\end{tabular}
}
\end{table}
%================================================
\subsubsection{Micro Ablation III: Optimization Dynamics (DWS Timing)}
Finally, we evaluate the critical timing of the information bottleneck penalty within the joint optimization (Table \ref{tab:dws_micro}). While the overall training is governed by the macroscopic three-stage DWS strategy, the exact timing to trigger the KL divergence penalty ($\beta(t)$) is the most sensitive and decisive factor in navigating the perception-distortion tradeoff. Therefore, this micro-ablation explicitly isolates the grace period of $\beta(t)$. Aggressively penalizing the KL divergence too early ($T_{warmup}=0$) forces premature information compression, which disrupts the establishment of cross-modal spatial mappings (PSNR drops to 28.58 dB). Conversely, delaying it too long ($T_{warmup}=30$) allows SAR speckle noise to deeply embed into the optical features. Furthermore, employing a hard truncation ($\beta(t)$ step jump) instead of smooth annealing leads to suboptimal convergence. Setting an exact 15-epoch grace period ($T_{warmup}=15$) ensures that the network establishes solid spatial alignment before physically truncating unstructured noise, achieving the optimal quantitative balance.
% --- TABLE 7: Cross-Corpus Adaptability with Cloud Ranges ---
\begin{table}[t]
\centering
\caption{Quantitative Performance on the LuojiaSET-OSFCR Dataset Across Different Cloud Coverage Levels. Best results are highlighted in \textbf{bold}.}
\label{tab:cross_sensor}
\renewcommand{\arraystretch}{1.3} 
\begin{tabular}{c l c c c c}
\toprule
\textbf{Cloud} & \textbf{Method} & \textbf{PSNR $\uparrow$} & \textbf{SSIM $\uparrow$} & \textbf{SAM $\downarrow$} & \textbf{MAE $\downarrow$} \\
\midrule
\multirow{3}{*}{0--20\%} 
& Cloudy   & 32.87 & 0.9473 & 2.77 & 0.0203 \\
& Baseline & 34.68 & 0.9602 & 2.66 & 0.0147 \\
& \textbf{IB-HFN} & \textbf{35.98}\rlap{~$\uparrow$} & \textbf{0.9628}\rlap{~$\uparrow$} & \textbf{2.07}\rlap{~$\downarrow$} & \textbf{0.0120}\rlap{~$\downarrow$} \\
\midrule
\multirow{3}{*}{20--40\%} 
& Cloudy   & 20.39 & 0.8159 & 5.67 & 0.0623 \\
& Baseline & 33.19 & 0.9216 & 2.97 & 0.0159 \\
& \textbf{IB-HFN} & \textbf{33.40}\rlap{~$\uparrow$} & \textbf{0.9267}\rlap{~$\uparrow$} & \textbf{2.74}\rlap{~$\downarrow$} & \textbf{0.0151}\rlap{~$\downarrow$} \\
\midrule
\multirow{3}{*}{40--60\%} 
& Cloudy   & 16.77 & 0.7221 & 7.54 & 0.1027 \\
& Baseline & 32.23 & 0.9034 & 3.08 & 0.0173 \\
& \textbf{IB-HFN} & \textbf{32.23}\rlap{~$\uparrow$} & \textbf{0.9104}\rlap{~$\uparrow$} & \textbf{2.97}\rlap{~$\downarrow$} & \textbf{0.0170}\rlap{~$\downarrow$} \\
\midrule
\multirow{3}{*}{60--80\%} 
& Cloudy   & 14.59 & 0.6588 & 9.32 & 0.1496 \\
& Baseline & 32.70 & 0.9070 & 3.37 & 0.0173 \\
& \textbf{IB-HFN} & \textbf{32.73}\rlap{~$\uparrow$} & \textbf{0.9144}\rlap{~$\uparrow$} & \textbf{3.12}\rlap{~$\downarrow$} & \textbf{0.0169}\rlap{~$\downarrow$} \\
\midrule
\multirow{3}{*}{80--100\%} 
& Cloudy   & 12.12 & 0.6101 & 12.31 & 0.2409 \\
& Baseline & 32.26 & 0.9086 & 3.71 & 0.0186 \\
& \textbf{IB-HFN} & \textbf{32.27}\rlap{~$\uparrow$} & \textbf{0.9172}\rlap{~$\uparrow$} & \textbf{3.45}\rlap{~$\downarrow$} & \textbf{0.0179}\rlap{~$\downarrow$} \\
\midrule
\multirow{3}{*}{\textbf{Overall}} 
& Cloudy   & 19.21 & 0.7489 & 7.55 & 0.1155 \\
& Baseline & 32.99 & 0.9196 & 3.16 & 0.0168 \\
& \textbf{IB-HFN} & \textbf{33.29}\rlap{~$\uparrow$} & \textbf{0.9258}\rlap{~$\uparrow$} & \textbf{2.88}\rlap{~$\downarrow$} & \textbf{0.0158}\rlap{~$\downarrow$} \\
\bottomrule
\end{tabular}
\end{table}
% ==============================================
% =================== 插入图 6：跨语料库极端云量视觉对比 ===================
\begin{figure}[t]
    \centering
    \includegraphics[width=1\linewidth]{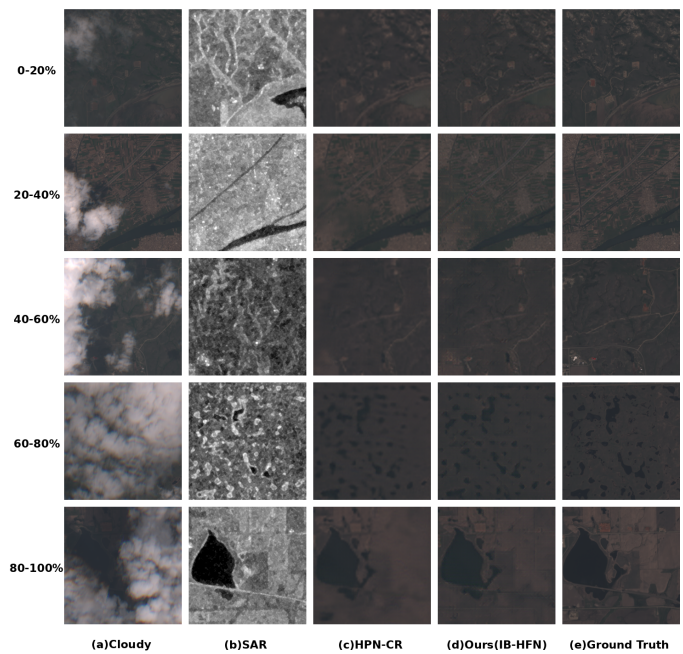} % 你的跨语料库测试对比图
     \vspace{-0.3cm}
    \caption{Qualitative comparison of cloud removal results on the LuojiaSET-OSFCR benchmark across five precise cloud coverage intervals (ranging from 0-20\% to 80-100\%). Compared to the baseline HPN-CR (c), which exhibits certain structural blurring and color shifts under extreme occlusion, the proposed IB-HFN (d) maintains better structural coherence and spectral fidelity, yielding results visually closer to the Ground Truth (e).}
    \label{fig:cross_corpus_visual} \vspace{-0.3cm}
\end{figure}
%====================
\subsubsection{Micro Ablation IV: Robustness to Data Volume Variations (Few-Shot Scenario)}
To further evaluate the data efficiency of the proposed framework, we conduct ablation studies by scaling down the training data volume. When we randomly sample only 30\% of the original training patches and evaluate on the same extreme test set, IB-HFN still maintains a competitive PSNR of 28.61 dB and a SAM of 7.54. This robustness indicates that our C-VIB layer effectively extracts deterministic geometric priors rather than strictly overfitting to massive paired data.
% =================== 插入图 8：可解释性特征图 ===================
\begin{figure}[t!]
    \centering
\includegraphics[width=1\linewidth, height=0.91\textheight, keepaspectratio]{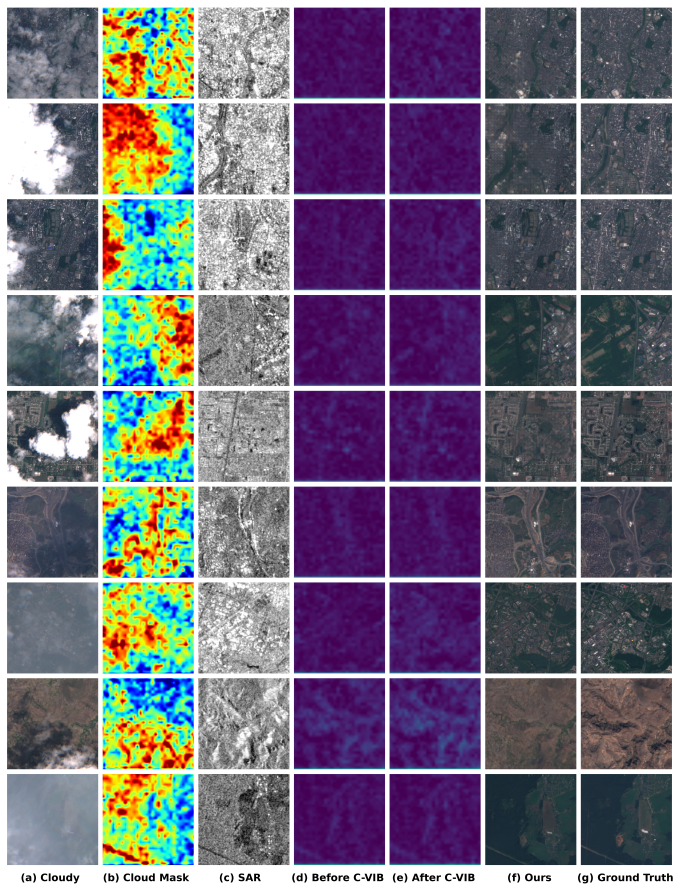}  % 可解释性图
     \vspace{-0.3cm}
    \caption{Intermediate feature visualization of the proposed SIBF module. As shown in (a) and (b), the visualized Cloud Mask accurately delineates both the thick cloud boundaries and their corresponding shadows. (Note: For intuitive human perception, column (b) visualizes the inversion of the network's internal clear-sky mask $M_{joint}$, with red regions highlighting heavy occlusion). Comparing the feature maps before (d) and after (e) the C-VIB layer, the unstructured SAR speckle noise from (c) is effectively suppressed, while the deterministic geometric structures are preserved. Guided by these purified features and masks, our IB-HFN reconstructs a high-fidelity cloud-free output (f) that closely aligns with the Ground Truth (g).}
    \label{fig:black_box}
    \vspace{-0.3cm}
\end{figure}
% ========================================================================

\subsection{Cross-Corpus Adaptability and Cloud Coverage Stress Test}
To rigorously verify the generalization capacity and learning robustness of the proposed IB-HFN framework, we extend our evaluation to the recently released LuojiaSET-OSFCR benchmark \cite{PAN2024258}. Although also derived from Sentinel satellites (featuring a spatial resolution of 10m), LuojiaSET-OSFCR provides a stringent stress test by sampling from 10 entirely independent, globally distributed, and non-overlapping Regions of Interest (ROIs) with highly complex land-cover diversity. The dataset comprises a total of 20,000 strictly co-registered image patch triplets, each with a fixed spatial dimension of $256 \times 256$ pixels. 

To comprehensively evaluate our model, we strictly partitioned the entire dataset into training, validation, and testing sets with an 8:1:1 ratio (comprising 16,000, 2,000, and 2,000 patches, respectively). More importantly, to conduct an extreme occlusion stress test, the testing set is further stratified into five precise cloud coverage intervals (0-20\%, 20-40\%, 40-60\%, 60-80\%, and 80-100\%, with approximately 400 test images per interval). We comprehensively retrain both the baseline and our IB-HFN on this new data distribution to assess their intrinsic capability to extract dataset-agnostic structural priors when confronted with novel geographic environments and varying degrees of extreme occlusion.

As reported in Table \ref{tab:cross_sensor} and visually corroborated in Fig. \ref{fig:cross_corpus_visual}, the original cloudy inputs suffer from severe degradation as cloud thickness increases (e.g., PSNR drastically drops from 32.86 dB at 0-20\% to 12.11 dB at 80-100\% occlusion). While the baseline HPN-CR demonstrates commendable robustness and effectively restores the images after retraining, it still exhibits slight spectral shifts and structural blurring when confronted with unfamiliar geographic distributions, as reflected by the SAM metric at extreme cloud coverage (80-100\%) and the visual degradation in Fig. \ref{fig:cross_corpus_visual}(c). Building upon this strong baseline, our IB-HFN yields consistent and measurable improvements across all cloud intervals. Notably, the overall SAM is further reduced from 3.1638 to 2.8773, alongside steady gains in PSNR and SSIM. This consistent performance gain indicates that the C-VIB layer effectively adapts to new spatial distributions. By learning dataset-agnostic physical priors, it successfully mitigates residual modality noise and bridges the geographic domain gap, proving its robust cross-corpus adaptability.
% =========================================================

\subsection{Opening the Black Box: Interpretability and Visualization}
To demystify the internal physical mechanism of IB-HFN, we visualize the intermediate feature maps in Fig. \ref{fig:black_box}. The Local-Global Gate successfully predicts accurate soft masks ($M_{joint}$) that delineate both the thick cloud boundaries and their corresponding cloud shadows. It is important to note that while the algorithm intrinsically computes a clear-sky mask ($M_{joint} \to 1$ for unclouded regions) to safely route optical textures, we explicitly visualize its inverse as a ``Cloud Mask'' (Fig. \ref{fig:black_box}(b)) to better align with human visual intuition, where high-heat (red) areas explicitly represent dense cloud occlusion. More importantly, visualizing channel responses before and after the C-VIB layer (Fig. \ref{fig:black_box}(d) vs. \ref{fig:black_box}(e)) reveals that channels encoding SAR granular noise are suppressed, while channels encoding urban buildings and land boundaries exhibit preserved structural backbones. This validates that our IB-HFN is rigorously filtering and sharpening from a physical perspective.
% ===============================Conclusion=========================================
\section{Conclusion}
\label{sec:con}

In this paper, we proposed IB-HFN, an information bottleneck driven framework for SAR assisted optical cloud removal. The key motivation is to suppress heterogeneous SAR speckle noise while preserving high frequency optical textures in clear sky regions. To this end, IB-HFN adopts a dual stream backbone to reduce early cross modal contamination and introduces a Spatial Information Bottleneck Fusion module for deep feature decoupling. Specifically, Channel-wise VIB (C-VIB) compresses SAR features to filter noisy responses, while a Local-Global Gate and a Dirac-initialized skip connection preserve reliable optical details. We further designed a joint optimization strategy with Dynamic Weight Scheduling, which coordinates feature level bottleneck regularization and image level high fidelity constraints to alleviate over smoothing and spectral distortion. Experiments on SEN12MS-CR under challenging spatio temporal splits demonstrate that IB-HFN achieves superior reconstruction quality, with 29.12 dB PSNR and 7.15 SAM, outperforming existing methods in structural preservation and spectral fidelity.

Despite its effectiveness, IB-HFN still has two limitations. First, the VGG based contrastive loss and KL regularization increase the training cost and memory usage. Second, the current framework relies on strictly co registered SAR optical image pairs, which may be difficult to obtain in rapid response scenarios. Future work will explore lightweight architectures, knowledge distillation, and self supervised information bottleneck learning to improve efficiency and reduce the dependence on paired training data.

\bibliographystyle{IEEEtran}
\bibliography{ref}

\end{document}